\documentclass[sigconf, screen]{acmart}
\AtBeginDocument{%
  \providecommand\BibTeX{{%
    \normalfont B\kern-0.5em{\scshape i\kern-0.25em b}\kern-0.8em\TeX}}}

\copyrightyear{2023}
\acmYear{2023}
\setcopyright{cc}
\setcctype[4.0]{by}
\acmConference[WWW '23 Companion]{Companion Proceedings of the ACM Web Conference 2023}{April 30-May 4, 2023}{Austin, TX, USA}
\acmBooktitle{Companion Proceedings of the ACM Web Conference 2023 (WWW '23 Companion), April 30-May 4, 2023, Austin, TX, USA}
\acmDOI{10.1145/3543873.3587362}
\acmISBN{978-1-4503-9419-2/23/04}

\makeatletter

\usepackage{caption}
\usepackage{enumitem}
\usepackage{microtype}
\usepackage{bm}
\usepackage{soul}
\usepackage{wrapfig}
\usepackage{amsmath}
\usepackage{mathtools}
\usepackage{balance}
\usepackage[varqu]{zi4}
\usepackage[utf8x]{inputenc}
\usepackage{units}
\usepackage{setspace}
\usepackage{printlen} %

\definecolor{redOV}{RGB}{255, 235, 238}
\definecolor{redI}{RGB}{255, 205, 210}
\definecolor{redII}{RGB}{239, 154, 154}
\definecolor{redIII}{RGB}{229, 115, 115}
\definecolor{redIV}{RGB}{239, 83, 80}
\definecolor{redV}{RGB}{244, 67, 54}
\definecolor{redVI}{RGB}{229, 57, 53}
\definecolor{redVII}{RGB}{211, 47, 47}
\definecolor{redVIII}{RGB}{198, 40, 40}
\definecolor{redIX}{RGB}{183, 28, 28}
\definecolor{redAI}{RGB}{255, 138, 128}
\definecolor{redAII}{RGB}{255, 82, 82}
\definecolor{redAIV}{RGB}{255, 23, 68}
\definecolor{redAVII}{RGB}{213, 0, 0}

\definecolor{pinkOV}{RGB}{252, 228, 236}
\definecolor{pinkI}{RGB}{248, 187, 208}
\definecolor{pinkII}{RGB}{244, 143, 177}
\definecolor{pinkIII}{RGB}{240, 98, 146}
\definecolor{pinkIV}{RGB}{236, 64, 122}
\definecolor{pinkV}{RGB}{233, 30, 99}
\definecolor{pinkVI}{RGB}{216, 27, 96}
\definecolor{pinkVII}{RGB}{194, 24, 91}
\definecolor{pinkVIII}{RGB}{173, 20, 87}
\definecolor{pinkIX}{RGB}{136, 14, 79}
\definecolor{pinkAI}{RGB}{255, 128, 171}
\definecolor{pinkAII}{RGB}{255, 64, 129}
\definecolor{pinkAIV}{RGB}{245, 0, 87}
\definecolor{pinkAVII}{RGB}{197, 17, 98}

\definecolor{purpleOV}{RGB}{243, 229, 245}
\definecolor{purpleI}{RGB}{225, 190, 231}
\definecolor{purpleII}{RGB}{206, 147, 216}
\definecolor{purpleIII}{RGB}{186, 104, 200}
\definecolor{purpleIV}{RGB}{171, 71, 188}
\definecolor{purpleV}{RGB}{156, 39, 176}
\definecolor{purpleVI}{RGB}{142, 36, 170}
\definecolor{purpleVII}{RGB}{123, 31, 162}
\definecolor{purpleVIII}{RGB}{106, 27, 154}
\definecolor{purpleIX}{RGB}{74, 20, 140}
\definecolor{purpleAI}{RGB}{234, 128, 252}
\definecolor{purpleAII}{RGB}{224, 64, 251}
\definecolor{purpleAIV}{RGB}{213, 0, 249}
\definecolor{purpleAVII}{RGB}{170, 0, 255}

\definecolor{deeppurpleOV}{RGB}{237, 231, 246}
\definecolor{deeppurpleI}{RGB}{209, 196, 233}
\definecolor{deeppurpleII}{RGB}{179, 157, 219}
\definecolor{deeppurpleIII}{RGB}{149, 117, 205}
\definecolor{deeppurpleIV}{RGB}{126, 87, 194}
\definecolor{deeppurpleV}{RGB}{103, 58, 183}
\definecolor{deeppurpleVI}{RGB}{94, 53, 177}
\definecolor{deeppurpleVII}{RGB}{81, 45, 168}
\definecolor{deeppurpleVIII}{RGB}{69, 39, 160}
\definecolor{deeppurpleIX}{RGB}{49, 27, 146}
\definecolor{deeppurpleAI}{RGB}{179, 136, 255}
\definecolor{deeppurpleAII}{RGB}{124, 77, 255}
\definecolor{deeppurpleAIV}{RGB}{101, 31, 255}
\definecolor{deeppurpleAVII}{RGB}{98, 0, 234}

\definecolor{indigoOV}{RGB}{232, 234, 246}
\definecolor{indigoI}{RGB}{197, 202, 233}
\definecolor{indigoII}{RGB}{159, 168, 218}
\definecolor{indigoIII}{RGB}{121, 134, 203}
\definecolor{indigoIV}{RGB}{92, 107, 192}
\definecolor{indigoV}{RGB}{63, 81, 181}
\definecolor{indigoVI}{RGB}{57, 73, 171}
\definecolor{indigoVII}{RGB}{48, 63, 159}
\definecolor{indigoVIII}{RGB}{40, 53, 147}
\definecolor{indigoIX}{RGB}{26, 35, 126}
\definecolor{indigoAI}{RGB}{140, 158, 255}
\definecolor{indigoAII}{RGB}{83, 109, 254}
\definecolor{indigoAIV}{RGB}{61, 90, 254}
\definecolor{indigoAVII}{RGB}{48, 79, 254}

\definecolor{blueOV}{RGB}{227, 242, 253}
\definecolor{blueI}{RGB}{187, 222, 251}
\definecolor{blueII}{RGB}{144, 202, 249}
\definecolor{blueIII}{RGB}{100, 181, 246}
\definecolor{blueIV}{RGB}{66, 165, 245}
\definecolor{blueV}{RGB}{33, 150, 243}
\definecolor{blueVI}{RGB}{30, 136, 229}
\definecolor{blueVII}{RGB}{25, 118, 210}
\definecolor{blueVIII}{RGB}{21, 101, 192}
\definecolor{blueIX}{RGB}{13, 71, 161}
\definecolor{blueAI}{RGB}{130, 177, 255}
\definecolor{blueAII}{RGB}{68, 138, 255}
\definecolor{blueAIV}{RGB}{41, 121, 255}
\definecolor{blueAVII}{RGB}{41, 98, 255}

\definecolor{lightblueOV}{RGB}{225, 245, 254}
\definecolor{lightblueI}{RGB}{179, 229, 252}
\definecolor{lightblueII}{RGB}{129, 212, 250}
\definecolor{lightblueIII}{RGB}{79, 195, 247}
\definecolor{lightblueIV}{RGB}{41, 182, 246}
\definecolor{lightblueV}{RGB}{3, 169, 244}
\definecolor{lightblueVI}{RGB}{3, 155, 229}
\definecolor{lightblueVII}{RGB}{2, 136, 209}
\definecolor{lightblueVIII}{RGB}{2, 119, 189}
\definecolor{lightblueIX}{RGB}{1, 87, 155}
\definecolor{lightblueAI}{RGB}{128, 216, 255}
\definecolor{lightblueAII}{RGB}{64, 196, 255}
\definecolor{lightblueAIV}{RGB}{0, 176, 255}
\definecolor{lightblueAVII}{RGB}{0, 145, 234}

\definecolor{cyanOV}{RGB}{224, 247, 250}
\definecolor{cyanI}{RGB}{178, 235, 242}
\definecolor{cyanII}{RGB}{128, 222, 234}
\definecolor{cyanIII}{RGB}{77, 208, 225}
\definecolor{cyanIV}{RGB}{38, 198, 218}
\definecolor{cyanV}{RGB}{0, 188, 212}
\definecolor{cyanVI}{RGB}{0, 172, 193}
\definecolor{cyanVII}{RGB}{0, 151, 167}
\definecolor{cyanVIII}{RGB}{0, 131, 143}
\definecolor{cyanIX}{RGB}{0, 96, 100}
\definecolor{cyanAI}{RGB}{132, 255, 255}
\definecolor{cyanAII}{RGB}{24, 255, 255}
\definecolor{cyanAIV}{RGB}{0, 229, 255}
\definecolor{cyanAVII}{RGB}{0, 184, 212}

\definecolor{tealOV}{RGB}{224, 242, 241}
\definecolor{tealI}{RGB}{178, 223, 219}
\definecolor{tealII}{RGB}{128, 203, 196}
\definecolor{tealIII}{RGB}{77, 182, 172}
\definecolor{tealIV}{RGB}{38, 166, 154}
\definecolor{tealV}{RGB}{0, 150, 136}
\definecolor{tealVI}{RGB}{0, 137, 123}
\definecolor{tealVII}{RGB}{0, 121, 107}
\definecolor{tealVIII}{RGB}{0, 105, 92}
\definecolor{tealIX}{RGB}{0, 77, 64}
\definecolor{tealAI}{RGB}{167, 255, 235}
\definecolor{tealAII}{RGB}{100, 255, 218}
\definecolor{tealAIV}{RGB}{29, 233, 182}
\definecolor{tealAVII}{RGB}{0, 191, 165}

\definecolor{greenOV}{RGB}{232, 245, 233}
\definecolor{greenI}{RGB}{200, 230, 201}
\definecolor{greenII}{RGB}{165, 214, 167}
\definecolor{greenIII}{RGB}{129, 199, 132}
\definecolor{greenIV}{RGB}{102, 187, 106}
\definecolor{greenV}{RGB}{76, 175, 80}
\definecolor{greenVI}{RGB}{67, 160, 71}
\definecolor{greenVII}{RGB}{56, 142, 60}
\definecolor{greenVIII}{RGB}{46, 125, 50}
\definecolor{greenIX}{RGB}{27, 94, 32}
\definecolor{greenAI}{RGB}{185, 246, 202}
\definecolor{greenAII}{RGB}{105, 240, 174}
\definecolor{greenAIV}{RGB}{0, 230, 118}
\definecolor{greenAVII}{RGB}{0, 200, 83}

\definecolor{lightgreenOV}{RGB}{241, 248, 233}
\definecolor{lightgreenI}{RGB}{220, 237, 200}
\definecolor{lightgreenII}{RGB}{197, 225, 165}
\definecolor{lightgreenIII}{RGB}{174, 213, 129}
\definecolor{lightgreenIV}{RGB}{156, 204, 101}
\definecolor{lightgreenV}{RGB}{139, 195, 74}
\definecolor{lightgreenVI}{RGB}{124, 179, 66}
\definecolor{lightgreenVII}{RGB}{104, 159, 56}
\definecolor{lightgreenVIII}{RGB}{85, 139, 47}
\definecolor{lightgreenIX}{RGB}{51, 105, 30}
\definecolor{lightgreenAI}{RGB}{204, 255, 144}
\definecolor{lightgreenAII}{RGB}{178, 255, 89}
\definecolor{lightgreenAIV}{RGB}{118, 255, 3}
\definecolor{lightgreenAVII}{RGB}{100, 221, 23}

\definecolor{limeOV}{RGB}{249, 251, 231}
\definecolor{limeI}{RGB}{240, 244, 195}
\definecolor{limeII}{RGB}{230, 238, 156}
\definecolor{limeIII}{RGB}{220, 231, 117}
\definecolor{limeIV}{RGB}{212, 225, 87}
\definecolor{limeV}{RGB}{205, 220, 57}
\definecolor{limeVI}{RGB}{192, 202, 51}
\definecolor{limeVII}{RGB}{175, 180, 43}
\definecolor{limeVIII}{RGB}{158, 157, 36}
\definecolor{limeIX}{RGB}{130, 119, 23}
\definecolor{limeAI}{RGB}{244, 255, 129}
\definecolor{limeAII}{RGB}{238, 255, 65}
\definecolor{limeAIV}{RGB}{198, 255, 0}
\definecolor{limeAVII}{RGB}{174, 234, 0}

\definecolor{yellowOV}{RGB}{255, 253, 231}
\definecolor{yellowI}{RGB}{255, 249, 196}
\definecolor{yellowII}{RGB}{255, 245, 157}
\definecolor{yellowIII}{RGB}{255, 241, 118}
\definecolor{yellowIV}{RGB}{255, 238, 88}
\definecolor{yellowV}{RGB}{255, 235, 59}
\definecolor{yellowVI}{RGB}{253, 216, 53}
\definecolor{yellowVII}{RGB}{251, 192, 45}
\definecolor{yellowVIII}{RGB}{249, 168, 37}
\definecolor{yellowIX}{RGB}{245, 127, 23}
\definecolor{yellowAI}{RGB}{255, 255, 141}
\definecolor{yellowAII}{RGB}{255, 255, 0}
\definecolor{yellowAIV}{RGB}{255, 234, 0}
\definecolor{yellowAVII}{RGB}{255, 214, 0}

\definecolor{amberOV}{RGB}{255, 248, 225}
\definecolor{amberI}{RGB}{255, 236, 179}
\definecolor{amberII}{RGB}{255, 224, 130}
\definecolor{amberIII}{RGB}{255, 213, 79}
\definecolor{amberIV}{RGB}{255, 202, 40}
\definecolor{amberV}{RGB}{255, 193, 7}
\definecolor{amberVI}{RGB}{255, 179, 0}
\definecolor{amberVII}{RGB}{255, 160, 0}
\definecolor{amberVIII}{RGB}{255, 143, 0}
\definecolor{amberIX}{RGB}{255, 111, 0}
\definecolor{amberAI}{RGB}{255, 229, 127}
\definecolor{amberAII}{RGB}{255, 215, 64}
\definecolor{amberAIV}{RGB}{255, 196, 0}
\definecolor{amberAVII}{RGB}{255, 171, 0}

\definecolor{orangeOV}{RGB}{255, 243, 224}
\definecolor{orangeI}{RGB}{255, 224, 178}
\definecolor{orangeII}{RGB}{255, 204, 128}
\definecolor{orangeIII}{RGB}{255, 183, 77}
\definecolor{orangeIV}{RGB}{255, 167, 38}
\definecolor{orangeV}{RGB}{255, 152, 0}
\definecolor{orangeVI}{RGB}{251, 140, 0}
\definecolor{orangeVII}{RGB}{245, 124, 0}
\definecolor{orangeVIII}{RGB}{239, 108, 0}
\definecolor{orangeIX}{RGB}{230, 81, 0}
\definecolor{orangeAI}{RGB}{255, 209, 128}
\definecolor{orangeAII}{RGB}{255, 171, 64}
\definecolor{orangeAIV}{RGB}{255, 145, 0}
\definecolor{orangeAVII}{RGB}{255, 109, 0}

\definecolor{deeporangeOV}{RGB}{251, 233, 231}
\definecolor{deeporangeI}{RGB}{255, 204, 188}
\definecolor{deeporangeII}{RGB}{255, 171, 145}
\definecolor{deeporangeIII}{RGB}{255, 138, 101}
\definecolor{deeporangeIV}{RGB}{255, 112, 67}
\definecolor{deeporangeV}{RGB}{255, 87, 34}
\definecolor{deeporangeVI}{RGB}{244, 81, 30}
\definecolor{deeporangeVII}{RGB}{230, 74, 25}
\definecolor{deeporangeVIII}{RGB}{216, 67, 21}
\definecolor{deeporangeIX}{RGB}{191, 54, 12}
\definecolor{deeporangeAI}{RGB}{255, 158, 128}
\definecolor{deeporangeAII}{RGB}{255, 110, 64}
\definecolor{deeporangeAIV}{RGB}{255, 61, 0}
\definecolor{deeporangeAVII}{RGB}{221, 44, 0}

\definecolor{brownOV}{RGB}{239, 235, 233}
\definecolor{brownI}{RGB}{215, 204, 200}
\definecolor{brownII}{RGB}{188, 170, 164}
\definecolor{brownIII}{RGB}{161, 136, 127}
\definecolor{brownIV}{RGB}{141, 110, 99}
\definecolor{brownV}{RGB}{121, 85, 72}
\definecolor{brownVI}{RGB}{109, 76, 65}
\definecolor{brownVII}{RGB}{93, 64, 55}
\definecolor{brownVIII}{RGB}{78, 52, 46}
\definecolor{brownIX}{RGB}{62, 39, 35}

\definecolor{grayOV}{RGB}{250, 250, 250}
\definecolor{grayI}{RGB}{245, 245, 245}
\definecolor{grayII}{RGB}{238, 238, 238}
\definecolor{grayIII}{RGB}{224, 224, 224}
\definecolor{grayIV}{RGB}{189, 189, 189}
\definecolor{grayV}{RGB}{158, 158, 158}
\definecolor{grayVI}{RGB}{117, 117, 117}
\definecolor{grayVII}{RGB}{97, 97, 97}
\definecolor{grayVIII}{RGB}{66, 66, 66}
\definecolor{grayIX}{RGB}{33, 33, 33}

\definecolor{bluegrayOV}{RGB}{236, 239, 241}
\definecolor{bluegrayI}{RGB}{207, 216, 220}
\definecolor{bluegrayII}{RGB}{176, 190, 197}
\definecolor{bluegrayIII}{RGB}{144, 164, 174}
\definecolor{bluegrayIV}{RGB}{120, 144, 156}
\definecolor{bluegrayV}{RGB}{96, 125, 139}
\definecolor{bluegrayVI}{RGB}{84, 110, 122}
\definecolor{bluegrayVII}{RGB}{69, 90, 100}
\definecolor{bluegrayVIII}{RGB}{55, 71, 79}
\definecolor{bluegrayIX}{RGB}{38, 50, 56}
\definecolor{bluegrayX}{RGB}{17, 23, 26}

\definecolor{myACMBlue}{cmyk}{1,0.1,0,0.1}
\definecolor{myACMYellow}{cmyk}{0,0.16,1,0}
\definecolor{myACMOrange}{cmyk}{0,0.42,1,0.01}
\definecolor{myACMRed}{cmyk}{0,0.90,0.86,0}
\definecolor{myACMLightBlue}{cmyk}{0.49,0.01,0,0}
\definecolor{myACMGreen}{cmyk}{0.20,0,1,0.19}
\definecolor{myACMPurple}{cmyk}{0.55,1,0,0.15}
\definecolor{myACMDarkBlue}{cmyk}{1,0.58,0,0.21}

\newcommand{\link}[1]{{\href{#1}{\color{blueVI}\textbf{\texttt{#1}}}}}
\newcommand{\linkhere}[2]{{\href{#1}{\color{blueVI}\textbf{#2}}}}

\newcommand{\figpart}[1]{\textcolor{myACMPurple}{#1}}
\newcommand{\mypar}[1]{\vspace{3px}\noindent\textbf{#1}\phantom{.}}

\newcommand{\headerspace}[0]{\vspace{-3pt}}
\newcommand{\headerspacebottom}[0]{\vspace{0pt}}

\newcommand{\tool}{WebSHAP}
\newcommand{\app}{Loan Explainer}

\newcommand*{\vcenteredhbox}[1]{\begingroup\setbox0=\hbox{#1}\parbox{\wd0}{\box0}\endgroup}

\begin{document}

\title{\tool{}: Towards Explaining Any Machine Learning Models Anywhere}

\author{Zijie J. Wang}
\orcid{0000-0003-4360-1423}
\email{jayw@gatech.edu}
\affiliation{%
\institution{Georgia Institute of Technology}
\city{Atlanta}
\state{Georgia}
\country{USA}
}

\author{Duen Horng Chau}
\orcid{0000-0001-9824-3323}
\email{polo@gatech.edu}
\affiliation{%
\institution{Georgia Institute of Technology}
\city{Atlanta}
\state{Georgia}
\country{USA}
}

\begin{abstract}
  As machine learning (ML) is increasingly integrated into our everyday Web experience, there is a call for transparent and explainable web-based ML.
  However, existing explainability techniques often require dedicated backend servers, which limit their usefulness as the Web community moves toward in-browser ML for lower latency and greater privacy.
  To address the pressing need for a client-side explainability solution, we present \tool{}, the first in-browser tool that adapts the state-of-the-art model-agnostic explainability technique SHAP to the Web environment.
  Our open-source tool is developed with modern Web technologies such as WebGL that leverage client-side hardware capabilities and make it easy to integrate into existing Web ML applications.
  We demonstrate \tool{} in a usage scenario of explaining ML-based loan approval decisions to loan applicants.
  Reflecting on our work, we discuss the opportunities and challenges for future research on transparent Web ML.
  \tool{} is available at \link{https://github.com/poloclub/webshap}.
\end{abstract}




\begin{teaserfigure}
  \centering
  \includegraphics[width=495pt]{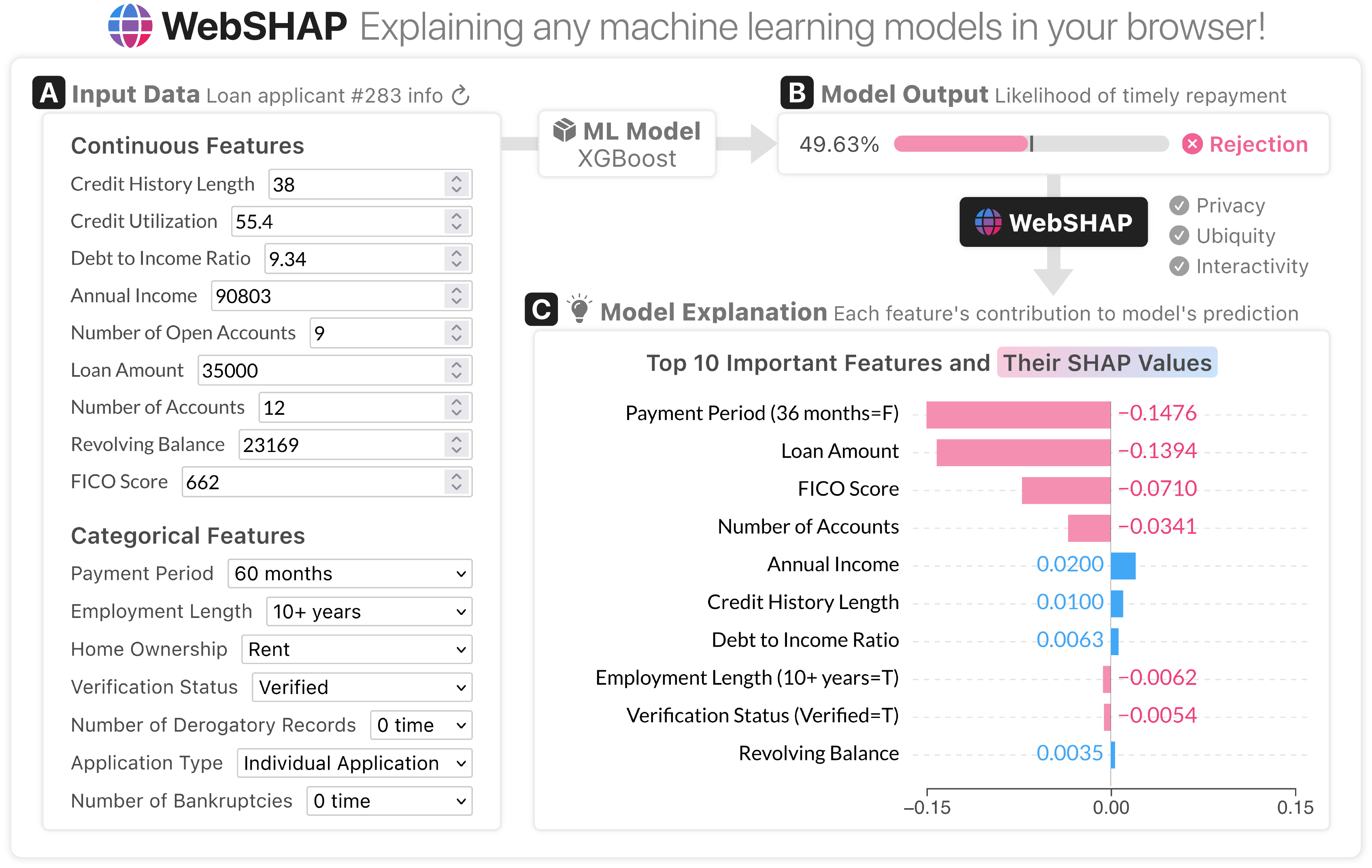}
  \vspace{-10pt}
  \caption{
    \tool{} is the first open-source tool for explaining any machine learning (ML) models in browsers.
    Adopting the state-of-the-art explainability technique Kernel SHAP to the Web, \tool{} offers \textit{private}, \textit{ubiquitous}, and \textit{interactive} explanations.
    For example, researchers can use \tool{} to build a client-side application that explains ML-based loan approval decisions to applicants, increasing their trust in ML models.
    \textbf{(A)} With this application, users can experiment with different feature inputs through the Web UI.
    \textbf{(B)} The application then updates the model prediction via in-browser inference.
    \textbf{(C)} \tool{} leverages modern Web technologies to compute feature importance in real time, delivering interactive and engaging explanations.
  }
  \Description{A screenshot of a tool Loan Explainer that is developed using \tool{} to explain a bank's loan approval decisions.}
  \label{fig:teaser}
\end{teaserfigure}

\maketitle

\headerspace{}
\section{Introduction}
\headerspacebottom{}

Machine Learning (ML) is now an integral part of our daily Web experiences, from social media feed ranking~\cite{jamaliHeteroMFRecommendationHeterogeneous2013}, fraud detection~\cite{panditNetprobeFastScalable2007}, to accessibility~\cite{qianGeneratingAccurateCaption2021}.
To enrich web applications with cutting-edge ML capabilities, there has been an increasing interest in client-side ML technologies.
For example, W3C has recently launched the Web ML Working Group to develop standards to enable in-browser ML inferences with access to client-side capabilities.
By offloading data processing to on-device hardware, users can enjoy ML-enriched web applications with lower latency and greater privacy.

Grounding in W3C's mission of ``ensuring the long-term growth of the Web'', the Web ML working group has adopted UNESCO AI ethical principles~\cite{unescoRecommendationEthicsArtificial2021} as the ethical principles for Web ML~\cite{fletcherEthicalPrinciplesWeb2022}.
A key principle is ``Transparency and Explainability,'' stating that Web ML developers should make their models explainable, inform users with reasons why an ML model makes certain decisions, and offer users means to contest and correct Ml-based decisions.
However, as ML models become increasingly complex, they can behave like black boxes, making it challenging to understand how they make predictions~\cite{wortmanvaughanHumancenteredAgendaIntelligible2022}.
This black-box challenge is even greater on the Web, as existing explainability techniques are computationally expensive and require dedicated servers~\cite[e.g.,][]{ribeiroWhyShouldTrust2016a,lundbergUnifiedApproachInterpreting2017a}.
How can we explain ML models on the Web and improve users' Web experience?

To tackle this pressing challenge on the Web, we present \textbf{\tool{}}~(\autoref{fig:teaser}), the first in-browser tool that offloads ML explainability to the client---empowering Web users to explain any ML models anywhere.
Our work makes the following key \textbf{contributions}:

\begin{itemize}[topsep=5pt, itemsep=0mm, parsep=1mm, leftmargin=9pt]
  \item \textbf{\tool{}, the first in-browser tool} that enables users to explain any ML predictions on the Web.
  Our tool adapts the state-of-the-art model-agnostic explainability technique Kernel SHAP~\cite{lundbergUnifiedApproachInterpreting2017a} to the Web environment.
  Kernel SHAP uses a game theoretic approach to compute feature importance for ML predictions.
  Developed with modern Web technologies, such as WebGL, WebSHAP harnesses client-side hardware capabilities to generate ML explanations with both privacy and low latency~(\autoref{sec:design}).

  \item \textbf{A usage scenario of explaining ML models} on the Web.
  We demonstrate \tool{}'s capabilities in a scenario where a bank develops a client-side application \app{}\footnote{\app{} is available at \link{https://poloclub.github.io/webshap/?model=tabular}, open sourced at \link{https://github.com/poloclub/webshap/tree/main/examples/demo}} to explain ML-based loan approval decisions to applicants~(\autoref{sec:usage}).
  We highlight the benefits of on-device explainability regarding \textit{privacy}, \textit{ubiquity}, and \textit{interactivity}~\cite{hortonLayerWiseDataFreeCNN2022}.
  Finally, we reflect on the opportunities and challenges for future research on transparent Web ML~(\autoref{sec:discussion}).

  \item \textbf{An open-source implementation} that lowers the barrier to applying explainability techniques to understand ML behaviors on the Web.
  We provide comprehensive documentation and examples to help developers easily adapt \tool{} to different Web environments.
  We develop \tool{} using TypeScript, a statically typed programming language that enables more maintainable and scalable code and allows users to easily integrate \tool{} into their Web ML applications~(\autoref{sec:design:implementation}).

\end{itemize}

\noindent In pursuit of the “View Source” ethos of the Web, \tool{} sheds light on transparent Web ML and provides a foundation for future researchers to develop trustworthy ML technologies on the Web. %
\headerspace{}
\section{W\lowercase{eb}SHAP in Action}
\label{sec:usage}
\headerspacebottom{}

We present a hypothetical usage scenario to demonstrate how \tool{} can help Giulia, a Web developer at a bank, to explain ML-based loan approval decisions to loan applicants such as Luca.

\mypar{Banks Explaining Loan Approval Decisions.}
Giulia develops a Web application \linkhere{https://poloclub.github.io/webshap/?model=tabular}{\app{}} that helps loan applicants understand how the bank uses ML to make loan application decisions~(\autoref{fig:teaser}).
Her bank trained an XGBoost classifier~\cite{chenXGBoostScalableTree2016a} on past data (we use LendingClub~\cite{LendingClubOnline2018} in this scenario).
This dataset contains 9 continuous features and 7 categorical features with a binary label indicating whether a person pays back their loan in time.
\app{} allows users to enter input features~(\autoref{fig:teaser}\figpart{A}) and observe how they affect the predicted loan approval likelihood~(\autoref{fig:teaser}\figpart{B}).

\mypar{Developing Explanable Web ML.}
To protect end-users' privacy, the bank requires the tool to fully run on the client side, allowing end-users to securely enter sensitive financial information.
To meet this goal, Giulia converts the XGBoost model from Python to \texttt{ONNX} format, an open and interoperable ML model representation~\cite{baiONNXOpenNeural2019}.
She then uses the \texttt{ONNX Web Runtime} to run the model inference in browsers with WebAssembly and Web Workers.
To help end-users make sense of the model's predictions, Giulia uses \tool{} to explain the model in browsers.
She installs \tool{} through \texttt{npm} with only one command, and integrates it into \app{} in under 10 minutes by (1) writing a JavaScript function to wrap the model inference code, (2) computing missing feature values (required by the Kernel SHAP algorithm) with training data median, (3) and passing the user-entered data to \tool{}.
Finally, Giulia visualizes the features' importance scores in a bar chart~(\autoref{fig:teaser}\figpart{C}).

\mypar{Loan Applicants Interpreting Web ML.}
Luca is a prospective loan applicant who uses \app{} to explore how banks use ML models for loan approval decisions.
He opens the tool in the browser on his tablet and clicks the refresh icon~\vcenteredhbox{\includegraphics[height=8pt]{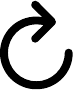}} several times until finding a preset applicant profile similar to his.
The tool updates the model's rejection decision and explanations this decision in real time.
By studying the feature importance bar chart and observing the changes in predictions after adjusting features, Luca discovers the significance of the~\vcenteredhbox{\includegraphics[height=9pt]{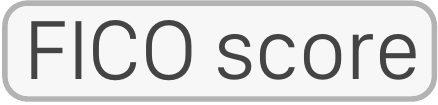}} in loan decision.
Impressed by the transparency of Giulia's bank, Luca decides to apply for a loan there after boosting his ~~\vcenteredhbox{\includegraphics[height=9pt]{figures/feature-fico}} above 700.
\headerspace{}
\section{System Design and Implementation}
\label{sec:design}
\headerspacebottom{}

\tool{} is the first tool for explaining any ML models on the Web.
To generate explanations with efficiency and privacy, it leverages the state-of-art explainability technique Kernel SHAP~(\autoref{sec:design:shap}) and harnesses modern Web technologies, such as WebGL, WebAssembly, and Web Workers~(\autoref{sec:design:technology}).
To help researchers and developers easily adopt \tool{}, we open-source our implementation and provide comprehensive documentation and tutorials~(\autoref{sec:design:implementation}).

\headerspace{}
\subsection{Adapting Kernel SHAP}
\label{sec:design:shap}
\headerspacebottom{}

SHAP is a state-of-art ML explainability framework popularized by \citet{lundbergUnifiedApproachInterpreting2017a}.
It uses the concept of Shapley values, originally applied in cooperative game theory for credit allocation~\cite{shapleyValueNpersonGames1953}, to calculate attribution scores for each feature for an individual ML prediction.
Shapley values are the average contribution of a player in all possible game coalitions.
In an ML context, players are input features, and game coalitions are permutations of input features compared to a baseline value.
Computing Shapley values is exponentially expensive, as one needs to iterate through all $2^M$ coalitions (permutations) of $M$ players (input features).

\newcommand{\lcolor}[1]{\textcolor{pinkIV}{#1}}
\newcommand{\rcolor}[1]{\textcolor{blueVII}{#1}}
\newcommand{\kcolor}[1]{\textcolor{orangeVI}{#1}}
\newcommand{\bcolor}[1]{\textcolor{black}{#1}}
\newcommand{\pcolor}[1]{\textcolor{tealVI}{#1}}

To more efficiently compute Shapley values, \citet{lundbergUnifiedApproachInterpreting2017a} introduce Kernel SHAP, a model-agnostic method to approximate the Shapley values of feature $x$ by solving a \lcolor{least squares problem $L$} through a \rcolor{linear regression $g$} with a \kcolor{kernel weight $\pi$}.
\begin{align}
\label{eq:shap}
    \begin{split}
    &\lcolor{L\left(\bcolor{f}, \rcolor{g}, \kcolor{\pi_{x}} \right)} = \sum_{z'\in{}Z}\left[\bcolor{f\left(h_x\left(z'\right)\right)} - \rcolor{g(z')}\right]^2 \kcolor{\pi_{x}\left(z'\right)} \\
    &\rcolor{g\left(z'\right)} = \pcolor{\phi_0} + \sum_{j=1}^M \pcolor{\phi_j} z_j'
    \hspace{12pt}
    \kcolor{\pi_{x}\left(z'\right)} = \frac{\left(M-1\right)}{\binom{M}{\left|z'\right|}\left|z'\right| \left(M - \left|z'\right|\right)}
    \end{split}
\end{align}
Here, $f$ is the ML model that we want to explain, and $f\left(h_x\left(z'\right)\right)$ is the model's predictions on the sampled data $h_x\left(z'\right)$, where each row contains features masked as missing~($z'_j=0$).
Users can specify the values to represent missing features through $h_x$, such as filling them with zeros, a subset of training data, or the median of training data.
The \rcolor{explanation model} $\rcolor{g\left(z'\right)}$ is a \rcolor{linear function} of binary variables $z' \in \{0, 1\}^M$, where $M$ is the number of input features.
The \kcolor{kernel} \kcolor{$\pi_{x}\left(z'\right)$} assigns a \kcolor{scalar weight} to each sampled instance $z'$ based on the number of non-missing features $\left|z'\right|$.
Finally, the \lcolor{least squares problem's} \pcolor{estimated solutions $\phi_j$} are the Shapley values.

We use Kernel SHAP as the explainability technique in \tool{} because it is state-of-the-art~\cite{hanWhichExplanationShould2022} and \textit{model-agnostic}.
This means with our tool, users can explain the predictions of \textit{any} ML model available on the Web, regardless of its architecture or implementation details.
Additionally, Kernel SHAP is the most favored explainability technique among ML practitioners according to a recent survey~\cite{krishnaDisagreementProblemExplainable2022}.
By using Kernel SHAP, we aim to make it easier for developers and researchers to adopt \tool{} and to provide them with the best explanations for their Web ML models.

\headerspace{}
\subsection{Optimizing for the Web}
\label{sec:design:technology}
\headerspacebottom{}

\mypar{Dataset Sampling.}
Solving the \lcolor{weighted least square problem} in \autoref{eq:shap} can be computationally challenging, as there are $2^M$ feature permutations ($|Z| = 2^M$).
\tool{} tackles this challenge by implementing a dataset sampling strategy as described in \cite{lundbergUnifiedApproachInterpreting2017a} and leveraging modern Web technologies.
First, \tool{} avoids sampling permutations for features with identical input and missing values, as their Shapley values will always be zero.
When dealing with input data with many features ($M > 30$), \tool{} does not sample all feature permutations: it prioritizes permutations with a large or small $|z'|$, as these instances have a larger \kcolor{kernel weight} \kcolor{$\pi_{x}\left(z'\right)$}, and thus provide more contribution to the \pcolor{solutions $\phi_j$}.

\mypar{Leveraging Modern Web Technologies.}
\tool{} employs the latest advancements in Web technologies and tooling to provide efficient and effective explanations of Web ML models.
For example, when solving the \lcolor{weighted least square problem}, \tool{} uses WebGL to accelerate matrix multiplications through \textit{TensorFlow.js}~\cite{smilkovTensorFlowJsMachine2019}, where matrices are stored as WebGL textures and matrix multiplication is implemented in a WebGL shader.
For instance, using the FireFox browser on a MacBook, \tool{} only takes about 600ms to multiply two matrices with dimensions of $2134 \times 2134$ through WebGL, a significant improvement from the 18 seconds it would take without WebGL.
Additionally, we provide examples that use Web Workers to run \tool{} in background threads to ensure that the Shapley value computation does not block the UI thread in browsers.
Finally, \tool{} is model-agnostic and capable of explaining any ML models available on the Web, including models compiled from non-Web languages.
One can even use \tool{} to explain an ML model running in a WebAssembly sandbox environment~(e.g., the ML models in \autoref{sec:usage} and in \autoref{sec:appendix}).

\headerspace{}
\subsection{Open-source and Easy to Use}
\label{sec:design:implementation}
\headerspacebottom{}

To help Web developers and researchers easily adopt \tool{}, we open source our implementation and design an API similar to the Kernel SHAP's Python implementation~\cite{lundbergUnifiedApproachInterpreting2017a}.
With \tool{}, explaining a Web ML model's prediction is as simple as two lines of code by passing a JavaScript prediction function, the missing feature values, and the data point.
Users also have the option to easily configure the number of feature permutations to sample ($|Z|$ in \autoref{eq:shap}).
Developed with TypeScript, our tool offers maintainable and scalable code, allowing users to easily extend and adapt it for their existing applications.
We provide detailed documentation and tutorials.\footnote{\tool{} documentation: \link{https://poloclub.github.io/webshap/doc}}
We publish \tool{} in the popular Web package repository \texttt{npm} Registry.\footnote{\tool{} \texttt{npm} repository: \link{https://www.npmjs.com/package/webshap}}
Users can easily install our tool and use it in both browser and \textit{Node.js}~\cite{dahlNodeJsOpensource2009} environments. %
\vspace{-1pt}
\section{Related Work}
\headerspacebottom{}

\mypar{Model-agnostic Explanation Methods.}
Researchers have proposed a wide array of model-agnostic explanation techniques~\citep[e.g.,][]{ribeiroWhyShouldTrust2016a,ribeiroAnchorsHighPrecision2018,lundbergUnifiedApproachInterpreting2017a}.
Given a trained ML model and a data point, these techniques aim to explain how different features contribute to the model's prediction.
Users can apply these techniques to any model class.
A recent survey with ML practitioners~\cite{krishnaDisagreementProblemExplainable2022} shows Kernel SHAP~\cite{lundbergUnifiedApproachInterpreting2017a}, which approximates feature attributions using a game theoretic approach, is the most favored technique.
Based on Kernel SHAP, researchers have proposed methods such as SAGE~\cite{covertUnderstandingGlobalFeature2020} for estimating global feature importance and \texttt{shapr}~\cite{aasExplainingIndividualPredictions2021} for models with many dependent features.
Advancing these related tools, our work is the first adaptation of Kernel SHAP for the Web.

\mypar{Explainable ML on the Web.}
The Web is a popular platform for explainable ML tools.
To help \textit{ML novices} learn about the inner workings of modern ML technologies, researchers develop Web-based visualization tools to interactively explain how different ML models work, such as GAN Lab~\cite{kahngGANLabUnderstanding2019a} and CNN Explainer~\cite{wangCNNExplainerLearning2020a}.
Researchers have also built web-based visual analytics tools to empower \textit{ML experts} to interpret their models~\cite[e.g.,][]{tenneyLanguageInterpretabilityTool2020,wexlerWhatIfToolInteractive2019,wangTimberTrekExploringCurating2022a}.
However, these tools often require dedicated backend servers to run ML models.
More recently, there is a growing number of explainability tools that can run entirely in the user’s browser.
For example, with pre-computation, Microscope~\cite{schubertOpenAIMicroscope2020} allows users to analyze neuron representations in their browsers.
GAM Changer~\cite{wangInterpretabilityThenWhat2022a}, a web-based tool to help users vet and fix the inherently interpretable Generalized Additive Models by running mode inference with WebAssembly.
In contrast, WebSHAP does not need backend servers or pre-computation, providing complete in-browser model explanations for any model class. %
\headerspace{}
\vspace{-10pt}
\section{Discussion and Future Work}
\label{sec:discussion}
\headerspacebottom{}

Reflecting on our development of \tool{}, we highlight the advantages and limitations of transparent and explainable Web ML.

\mypar{Advantages and Opportunities.}
The key benefits of enabling ML explainability on the Web are \textit{privacy}, \textit{ubiquity}, and \textit{interactivity}.
\tool{} empowers users to interpret Web ML models directly on their devices, keeping sensitive model inputs secure (e.g., financial and medical information).
As the Web is ubiquitous, users can use \tool{} on their computers, tablets, phones, and even IoT devices (e.g., smart refrigerators).
Using the Web as a platform, \tool{} makes it easier for developers to deploy explainable ML systems and enable user interactions.
\textbf{Future research opportunities} include:
\begin{itemize}[topsep=3pt, itemsep=0mm, parsep=2pt, leftmargin=9pt]
    \item \textbf{Enhancing \tool{} with new Web APIs} such as Service Worker for offline explainability, WebSocket for collaborative interpretations, and Web Crypto for verifiable explanations.

    \item \textbf{Integrating \tool{} directly into browsers} such as through the Web Inspector tools. It will allow users to easily view and interpret any ML models running on a Web page.

    \item \textbf{Developing web-based interactive visualization tools} to help end-users easily digest model explanations.
\end{itemize}

\setlength{\columnsep}{6pt}%
\setlength{\intextsep}{0pt}%
\begin{wrapfigure}{R}{0.17\textwidth}
  \vspace{-1pt}
  \centering
  \includegraphics[width=0.17\textwidth]{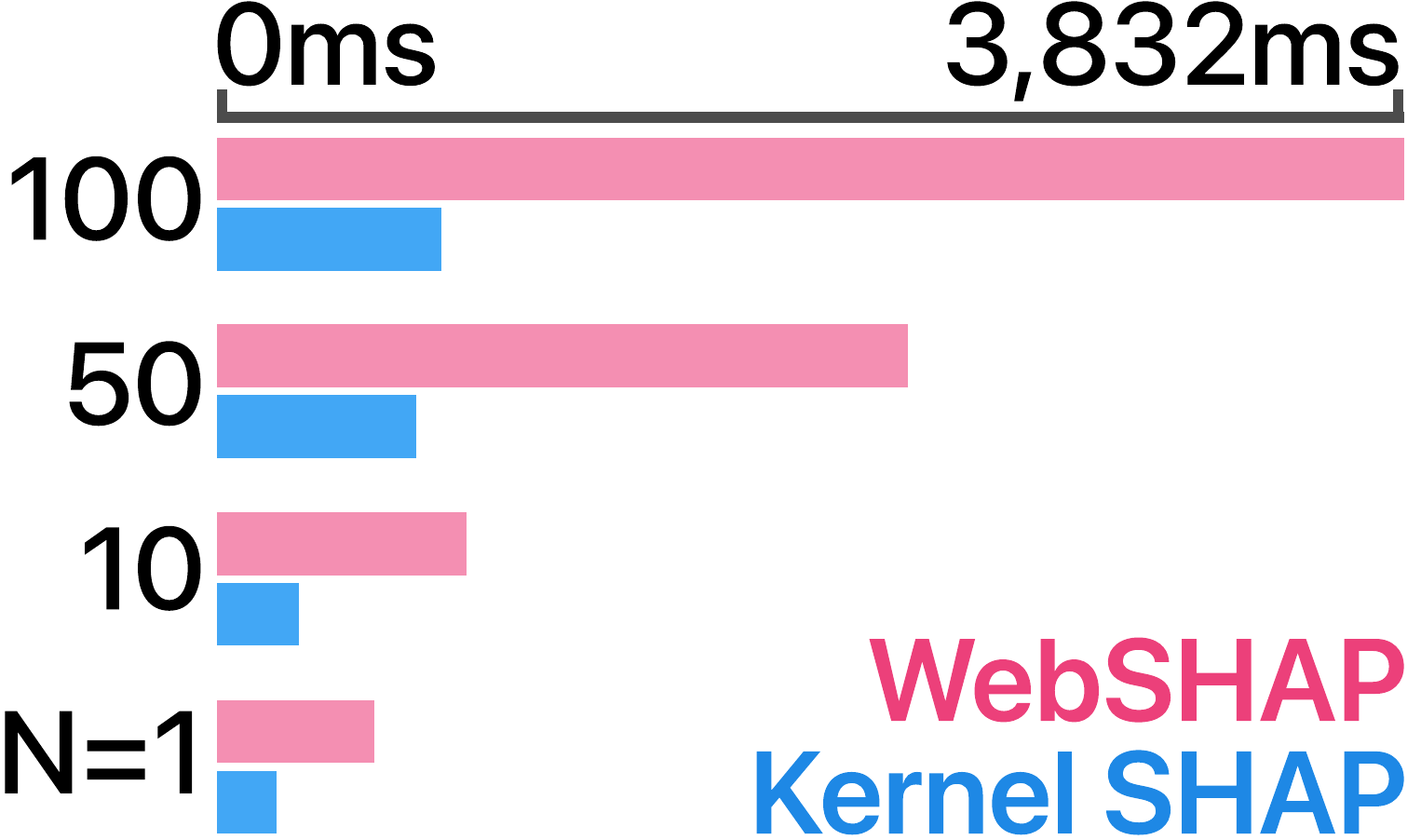}
\end{wrapfigure}
\mypar{Limitations and Challenges.}
We first acknowledge the limitations of using a post-hoc explainability technique, as it can produce inaccurate and unstable explanations~\cite{rudinStopExplainingBlack2019}.
Also, developing explainable ML models for the Web faces unique challenges, including limited computation resources in browsers, varying capacities among edge devices, and a lack of established Web ML APIs and libraries.
With the ML model in~\autoref{sec:usage}, we compare average SHAP computation times between \textcolor{pinkV}{\tool{}} and \textcolor{blueVI}{Kernel SHAP (Python)} across different background data size $N$ on a 64GB RAM MacBook (see right).
\textcolor{pinkV}{\tool{}} is slower than \textcolor{blueVI}{Kernel SHAP} especially when $N$ is large, and the main factor is the XGBoost inference time difference.

\headerspace{}
\section{Conclusion}
\headerspacebottom{}
We present \tool{}, an in-browser, open-source explainability library for Web ML.
Our tool adapts Kernel SHAP and leverages modern Web technologies for easy integration into existing Web ML applications.
To demonstrate its potential, we present a usage scenario demonstrating real-time explanation of a web-based loan approval prediction model.
In pursuit of the ``View Source'' ethos of the Web, we aim for \tool{} to be a stepping stone towards transparent, explainable, and trustworthy ML on the Web. %
\begin{acks}
  This work was supported in part by a J.P. Morgan PhD Fellowship, Apple Scholars in AI/ML PhD fellowship, and DARPA GARD.
\end{acks}

\begin{spacing}{0.971}
\bibliographystyle{ACM-Reference-Format}
\bibliography{23-webshap}
\end{spacing}

\appendix
\clearpage{}

\onecolumn
\setcounter{figure}{0}
\renewcommand{\thetable}{S\arabic{table}}
\renewcommand{\thefigure}{S\arabic{figure}}

\section{Supplementary Examples}
\label{sec:appendix}

\vspace{5pt}

\setlength{\belowcaptionskip}{0pt}
\setlength{\abovecaptionskip}{1pt}
\begin{figure*}[!htb]
  \includegraphics[width=426pt]{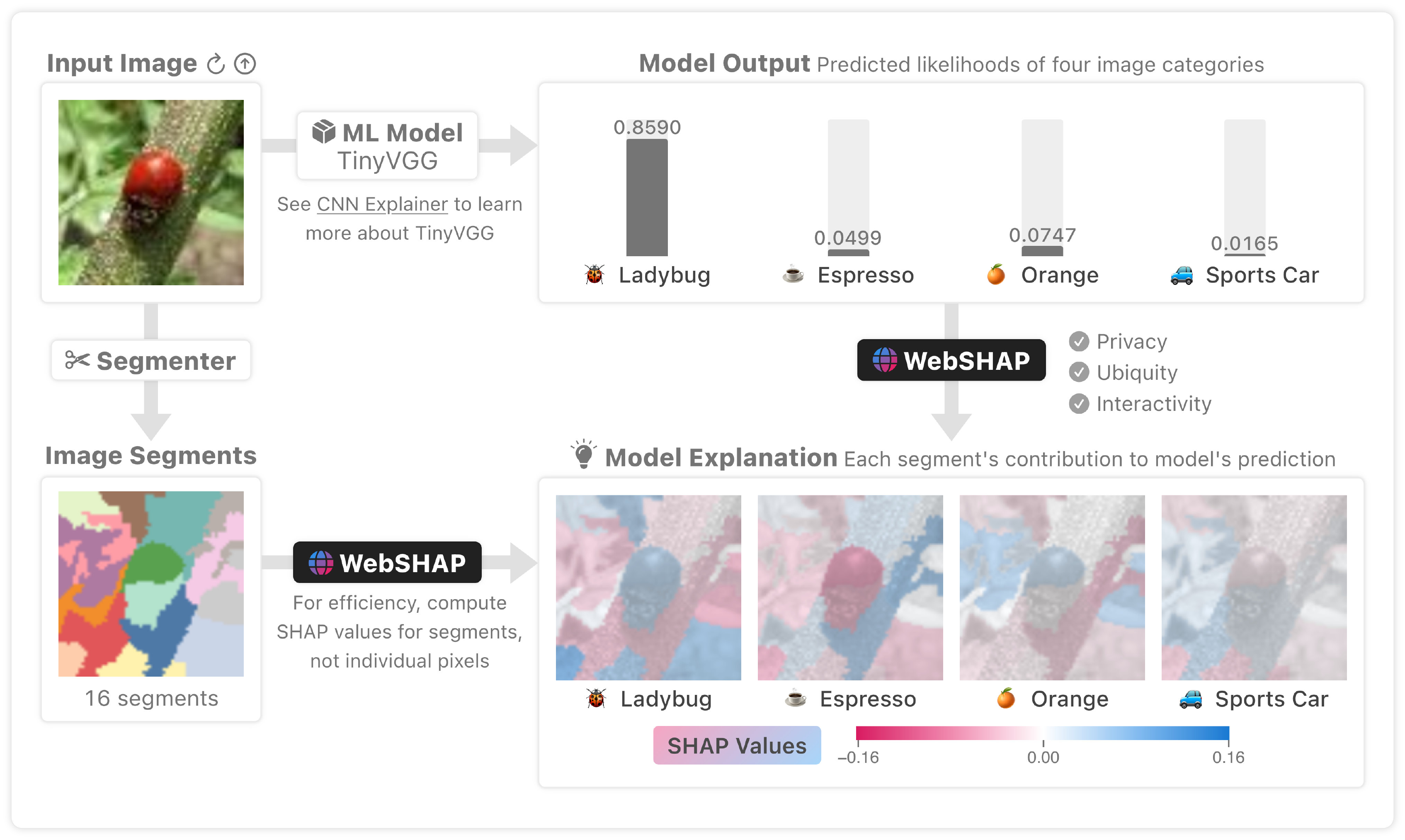}
  \caption{
    In this example, we use \tool{} to explain how a TinyVGG multi-class image classifier~\cite{wangCNNExplainerLearning2020a} makes predictions.
    We compute the SHAP values of 16 image segments and then visualize them using a diverging color scale overlay.
    Visualizations show that the model learns to capture the ladybug's shape and color.
    All components---image classifier, image segmenter, and \tool{}---run completely in the browser.
    This example is accessible at \link{https://poloclub.github.io/webshap/?model=image}.
  }
  \Description{
    A screenshot of a Web app that uses WebSHAP to explain an image classifier.
  }
  \label{fig:appendix-image}
\end{figure*}

\vspace{8pt}

\begin{figure*}[!htb]
  \includegraphics[width=426pt]{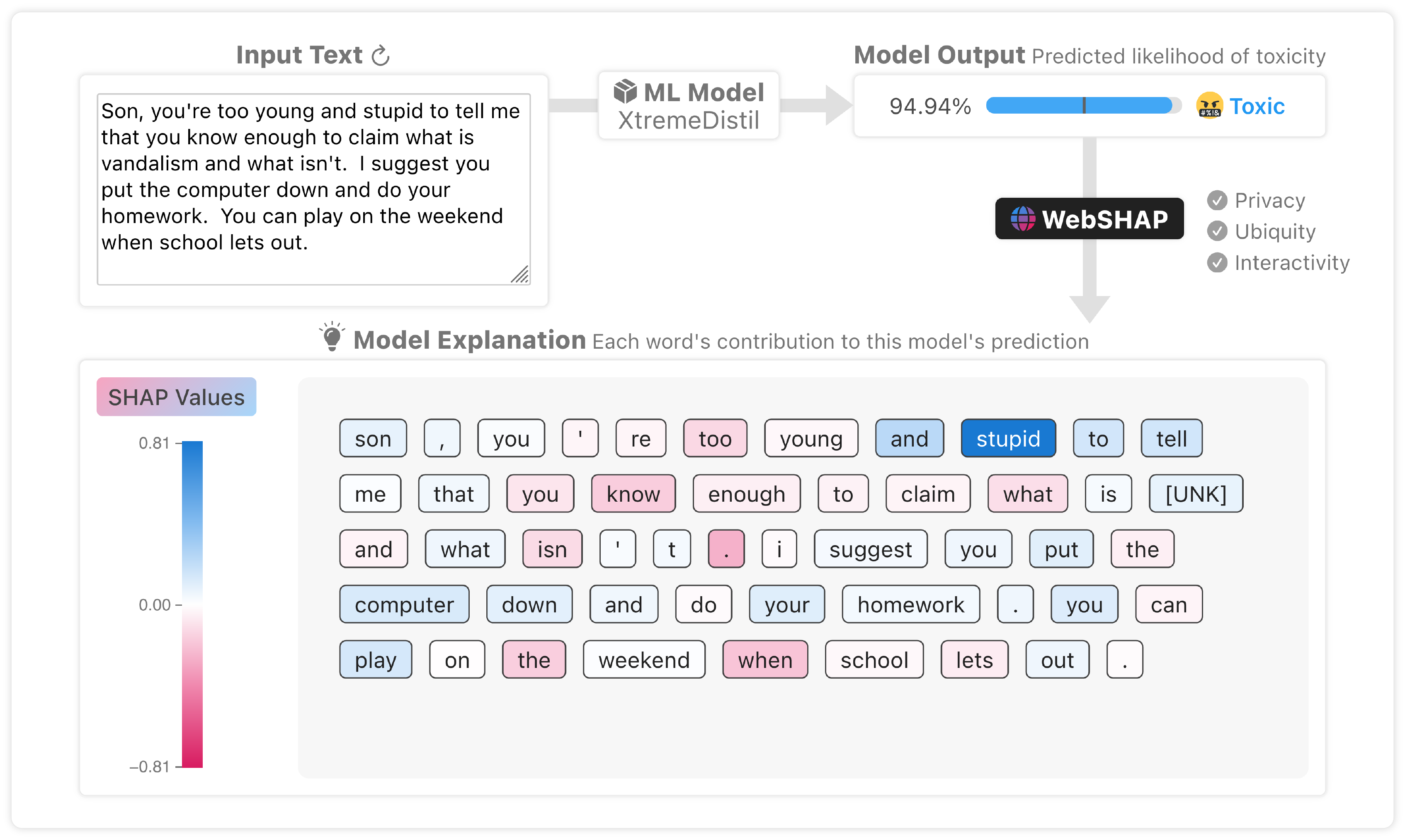}
  \caption{
    We use \tool{} to explain predictions of a transformer-based text toxicity detector.
    We compute the SHAP values for individual text tokens and visualize them using a diverging color scale applied to each token's background color.
    It shows negative words such as ``stupid'' and ``down'' contribute significantly to the toxicity prediction.
    The tokenizer, toxicity detector, and \tool{} run entirely on the client-side.
    This example is accessible at \link{https://poloclub.github.io/webshap/?model=text}.
  }
  \Description{
    A screenshot of a Web app that uses WebSHAP to explain a text toxicity detector.
  }
  \label{fig:appendix-text}
\end{figure*}
\setlength{\belowcaptionskip}{0pt}
\setlength{\abovecaptionskip}{12pt} 
\end{document}